\documentclass[10pt,twocolumn,letterpaper]{article}

\usepackage{cvpr}
\usepackage{times}
\usepackage{epsfig}
\usepackage{graphicx}
\usepackage{amsmath}
\usepackage{amssymb}

\usepackage{algorithm}
\usepackage{algorithmic}

\usepackage[breaklinks=true,bookmarks=false]{hyperref}

\cvprfinalcopy 

\def\cvprPaperID{****} 

\begin{document}
\title{On Isometry Robustness of Deep 3D Point Cloud Models\\ under Adversarial Attacks}
\author{Yue Zhao$^1$, Yuwei Wu$^1$, Caihua Chen$^2$\footnotemark[1], Andrew Lim$^1$\\
$^1$Department of Industrial Systems Engineering and Management, National University of Singapore\\
$^2$School of Management and Engineering, Nanjing University\\
{\tt\small \{yuezhao, ywwu\}@u.nus.edu}, 
{\tt\small chchen@nju.edu.cn}, 
{\tt\small isealim@nus.edu.sg}
}

\maketitle

\renewcommand{\thefootnote}{\fnsymbol{footnote}}
\footnotetext[1]{Corresponding author.}
\begin{abstract}
 While deep learning in 3D domain has achieved revolutionary performance in many tasks, the robustness of these models has not been sufficiently studied or explored. Regarding the 3D adversarial samples, most existing works focus on manipulation of local points, which may fail to invoke the global geometry properties, like robustness under linear projection that preserves the Euclidean distance, i.e., isometry. In this work, we show that existing state-of-the-art deep 3D models are extremely vulnerable to isometry transformations. Armed with the Thompson Sampling, we develop a black-box attack with success rate over 95\% on ModelNet40 data set. Incorporating with the Restricted Isometry Property, we propose a novel framework of white-box attack on top of spectral norm based perturbation. In contrast to previous works, our adversarial samples are experimentally shown to be strongly transferable. Evaluated on a sequence of prevailing 3D models, our white-box attack achieves success rates from 98.88\% to 100\%. It maintains a successful attack rate over 95\% even within an imperceptible rotation range $[\pm 2.81^{\circ}]$.  
\end{abstract}

\section{Introduction}

Recently deep learning in 3D data sets has increasingly raised attentions due to its huge potential in real-world applications such as self-driving cars \cite{geiger2012we,chen2017multi}, augmented reality \cite{park2008multiple} and medical image analysis \cite{milletari2016v}.  Although techniques in 2D deep learning like multi-view approaches \cite{su2015multi,qi2016volumetric} can be adopted for tackling 3D problems, it is hard to extend these approaches to some 3D tasks, for example point classification. To alleviate this difficulty, the authors of \cite{qi2017pointnet} proposed a novel deep neural network PointNet that directly consumes 3D point cloud data. With the proved invariance to order change, it performs well on point cloud classification and segmentation tasks. After this pioneering work, there is a surge of researches \cite{qi2017pointnet++,li2018so,li2018pointcnn,wang2019dynamic,liu2019relation} on 3D neural networks that process directly on point clouds and achieve promising results. 

\begin{figure}
\centering
\includegraphics[width=1\linewidth]{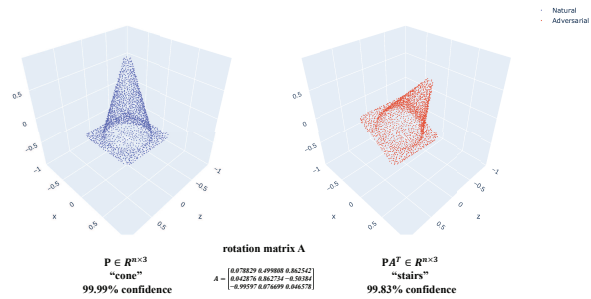}
\caption{Demonstration of Thompson Sampling Isometry attack on PointNet model \cite{qi2017pointnet} trained by ModelNet40 data \cite{wu20153d}. Such classification error may cause safety hazard for autonomous vehicles, since a rotated traffic cone is commonly seen in real world scenarios.}\label{fig_tsi}
\end{figure}

While deep 3D models have gained remarkable success, the invariance or robustness to isometry transformation, a fundamental concern in computer vision \cite{elad2003bending}, has not raised enough attention. It is also important to note that isometry represents a large class of geometry transformations including translation, rotation, and reflection as the most common examples.
A misclassification case of a highway cone under our black-box attack is illustrated in Figure \ref{fig_tsi}. To address this issue, in this paper we design a novel block-box attack framework based on Thompson Sampling (TS)  \cite{russo2018tutorial} and evaluate the robustness of the existing state-of-art deep 3D models (PointNet \cite{qi2017pointnet}, PointNet++ \cite{qi2017pointnet++}, DG-CNN \cite{wang2019dynamic} and  RS-CNN \cite{liu2019relation})  for classification problems under linear isometry. Specifically, in the black box TS is used to generate rotation matrix parameterized by its rotation angle, and the resulted rotation transformation can be applied to nature samples. The adoption of TS is motivated by the basic belief that the attack success rate varies in different rotation ranges. In order to generate successful attacks, a natural idea is to
sample with probability to rotation angles. Compared to random sampling, TS is more efficient and explainable due to its ability to adaptively learn the prior probability of parameters. With the TS techniques, our block box attack can achieve success rates more than 95.66\% on ModelNet40 data over all evaluated models by at most 10 times of iterative sampling. Heat map of attack success rates in different regions suggests certain rotation ranges are more likely to fool models, which confirms our insights. We also show links between our attacks and efficient 3D data augmentation subsequently.


Beyond isometry, we hope a deep 3D model can be robust under slightly perturbed isometry, since human beings have the capability to easily recognize a stretched 3D object from the natural one. This idea naturally leads to the Restricted Isometry Property (RIP) matrix \cite{candes2005decoding}, where RIP is a nice relaxation of isometry property extensively investigated in the compressed sensing field. In fact, a RIP matrix is coordinated with a small positive number called Restricted Isometric Constant (RIC), as a measurement of its distance from an isometry. In the white-box attack, we start from an initial isometry and further add perturbation on it to transform point cloud. The motivation is to preserve isometry as much as possible while misleading the model. Although finding RIC of a given matrix $A$ is computational hard \cite{tillmann2013computational} in general, it is equivalent to compute the spectral norm of $A^TA-I$ in our case. Besides, spectral norm offers a fair and universal standard of perturbation level among different point cloud data sets, which is similar to the $L_p$ norm standard in image domain. On an array of models trained by ModelNet40 data \cite{wu20153d}, we achieve attack success rates more than 98.11\% while the spectral norms are less than 0.068. 

The contributions of this paper are fourfold:
\begin{itemize}
    \item We utilize Thompson Sampling to design an efficient black-box attack with zero loss in the sense of isometry. It achieves success rates over 95\% on ModelNet40 data set, which is the first work to fully reveal the vulnerability of deep 3D point cloud models under isometry transformations in black-box settings. To the best of our knowledge, this is the first black-box attack in 3D point cloud adversarial settings.  
    \item Inspired by the theoretical results of Restricted Isometry Property, we introduce a novel framework of white-box attack on point cloud with success rates more than 98.11\% over several state-of-the-art models. Even for imperceptible rotation angles in $[-\pi/64,\pi/64]$ ($\pm 2.81^{\circ}$), it achieves a success rate over 95\%. 
    \item In contrast to previous works, 3D adversarial samples generated by our algorithms have significantly strong transferability. This property further shows the vulnerability of current models under isometry projections. It also suggests huge potential of our approach in developing further black-box attacks.
    
    \item Our adversarial samples are pervasive in practice. For example, calibration in self-driving car system is sometimes inaccurate \cite{tesla}, thus the input signals are often subjected to small perturbations, which can be viewed as our adversarial samples. Hence, our approach might offer insights for improving robustness of real-world applications.

\end{itemize}
\renewcommand{\thefootnote}{\arabic{footnote}}
The rest of the paper is organized as follows: We start with related works in deep 3D point cloud models, adversarial attacks in 3D settings and data augmentation. Then we offer technique details of isometry characterization, TS and RIP in preliminaries. After preliminaries, our attack methods are proposed with the pseudo-code of the algorithms. In the end, we evaluate attack performance for on state-of-the-art models trained by two benchmark data sets\footnote{https://github.com/skywalker6174/3d-isometry-robust}. Furthermore, we give an analysis of the trade-off between accuracy and robustness of models from a data augmentation perspective.

\section{Related Work}

\noindent
\textbf{Deep 3D Point Cloud Models.}   
3D data sets are of intrinsic difference from images, thus current neural network designed for 2D domain might not be directly applied in certain 3D tasks. To address this issue, \cite{qi2017pointnet} proposed PointNet, a neural network which directly processes on point clouds. Due to its efficiency and invariance to order changes of points, it has been widely accepted and developed \cite{qi2017pointnet++,wang2019dynamic, liu2019relation}. A concept called the critical set, is introduced in \cite{qi2017pointnet}. It plays an important role in enhancing model robustness, whereas \cite{wicker2019robustness} showed in practice it might instead be a weakness exploited by attacks. PointNet also used a mini-network called T-net and regularization term to make the feature transformation matrix near orthogonal matrix, so that it could be robust against rigid transformations (isometry). However, \cite{liu2019relation} noticed that accuracy of PointNet reduced to around 40\% under rotations about Y axis by $90^{\circ}$ or $180^{\circ}$. We observe that the issue of isometry invariance has not been well explored in previous works. It will be our main focus in this work.\\

\noindent
\textbf{Adversarial Samples in 3D.}
In image domain, adversarial attacks \cite{goodfellow2014explaining,papernot2016distillation,kurakin2016adversarial,carlini2017towards}, along with the defense approaches \cite{madry2017towards,athalye2018obfuscated,zhang2019theoretically}, have been extensively studied. However, 3D point cloud data is essentially coordinates, which are completely different from pixel values, thereby it is necessary to design new approaches for adversarial attacks in 3D. By utilizing critical points, \cite{zheng2018pointCloud} assigned each point a score measuring its contribution to model training loss, then built saliency maps for attack. With an optimization approach, \cite{xiang2019generating} generated adversarial point clouds by perturbation under $L_p$ norm and adding points in local area. They also made use of critical points as the initial points for solving optimization problems in attack. \cite{wicker2019robustness} developed a pointwise occlusion attack by iteratively excluding points from point clouds. \cite{yang2019adversarial} proposed a pointwise gradient-based perturbation attack and a point attachment/detachment attack, as well as statistical tools to distinguish between adversarial samples and noisy data for defense. All the above papers in 3D adversarial learning develop attacks based on local information of individual points, whereas in this work we analyse the robustness invoking the global geometry structure of point clouds.\\  

\noindent
\textbf{On Data Augmentation.}
For image problems, data augmentations are well-studied to improve the performance of deep learning models. Common strategies can be roughly divided into traditional methods (cropping, reflecting, rotating, scaling, and translation) and learning-based methods \cite{DeVries2017,Moosavi-Dezfooli2016,Goodfellow2014}, see \cite{Shorten2019} for a recent survey. Data augmentation is not so well-addressed on point clouds. To our knowledge, simple strategies, such as random scaling, translation, and rotation, are most commonly used in recent studies \cite{qi2017pointnet,qi2017pointnet++,li2018pointcnn,Zhou2018}. In this paper we show that these strategies are far from satisfactory. Consequently, the study of data augmentation, along with the adversarial attack, may have great potential to improve both accuracy and robustness of deep 3D models.\\

\noindent
\textbf{3D Isometry Robustness.} The isometry robustness considered in the past are mainly involved with 3D object retrieval \cite{kazhdan2003rotation,tangelder2004survey} and matching problems \cite{bronstein2006efficient} in computer vision. Although they are different from the classification problem in this paper, we believe many strategies can be adopted and applied in deep learning settings. There are two possible ways of incorporation: 1) normalization (pose estimation), for example, the most prominent one based on PCA \cite{vranic2001tools} can transform data to certain orientation, which could neutralize the effect of rotations, 2) isometry invariant descriptors based on computer vision theory, such as \cite{kazhdan2003rotation,novotni20033d}, which can be embedded into the network architectures and enhance isometry robustness. Recently there are a few concurrent works \cite{poulenard2019effective,zhang2019rotation,sun2019srinet} utilizing those ideas. However, in general they could suffer from more complex structures, more computational time or less accuracy. Our work fills the research gap between those attempts and prevailing deep 3D models by illustrating the vulnerability of the latter and justifying the importance of the former. Furthermore, we hope our methods can be served as an effective tool to evaluate isometry robustness and offer insights for future works.

\section{Preliminaries}
\subsection{Linear isometry and Orthogonal matrix}\label{liso&Orth}\label{subsec:iso}
\noindent
\textbf{Linear Isometry.} 
An \emph{isometry} of $\mathbb{R}^n$ is a function $f:\mathbb{R}^n\to\mathbb{R}^n$ that preserves the Euclidean distance
\begin{equation}
\Vert f(x)-f(y)\Vert = \Vert x-y\Vert \quad \forall x,y\in\mathbb{R}^n
\end{equation}
Trivial examples of isometry include the identity transformation $f(x)=x$ and translation $f(x)=x+b$.  An isometry that fixes the origin is indeed a linear transformation on $\mathbb{R}^n$ \cite{triola2009special}.  Because the effect of translations can always be easily eliminated by normalization, in this paper we only consider the case of linear isometry, i.e., $f$ is a linear transformation that preserve the Euclidean distance. In the rest of this paper, we mean linear isometry when referring to isometry. \\

\noindent
\textbf{Orthogonal matrix.} 
An $n\times n$ matrix $A$ is said to be \emph{orthogonal} if 
\begin{equation}\label{orth}
A\in \mathrm{O}(n):=\{A\in \mathrm{GL}_n(\mathbb{R}):A^TA=I\}
\end{equation}
where $\mathrm{GL}_n(\mathbb{R})$ denotes the set of $n\times n$ invertible matrices of real numbers. It is easy to prove that $\mathrm{det}(A)=\pm 1,\forall A\in \mathrm{O}(n)$ . Let $\mathrm{Iso}(n)$ denote the set of isometries of $\mathbb{R}^n$, \cite{triola2009special} (Lemma1.7, Proposition 1.8) shows
\begin{equation}
    \mathrm{Iso}(n)=\mathrm{O}(n)
\end{equation}
Rotation matrices are defined as a subset of $\mathrm{O}(n)$
\begin{equation}
    \mathrm{SO}(n):=\{A\in \mathrm{O}(n): \mathrm{det}(A)=1\}
\end{equation}
A reflection matrix over a hyper-plane passing origin is defined by the householder matrix \cite{householder1958unitary}
\begin{equation}\label{ref}
    P = I-2vv^T
\end{equation}
where $v$ is the unit normal vector of the hyper-plane $\{x\in\mathbb{R}^n:v^Tx=0\}$.\\

\noindent
\textbf{Parameterization of linear isometry.} By Cartan–Dieudonné theorem any matrix in $\mathrm{O}(3)$ can be described as at most three reflections, and a two reflections case in $\mathbb{R}^3$ is equal to a rotation. To further represent a rotation, the Euler's rotation theorem claims that in $\mathbb{R}^3$ we can always represent a rotation matrix by three rotation matrices about $x,y,z$ axes 
\begin{equation}\label{Euler}
R = R_{\theta_x}R_{\theta_y}R_{\theta_z}
\end{equation}
where $R_{\theta_i}$ is the rotation matrix about axis-i by angle $\theta_i\in[-\pi,\pi]$. Hence any rotation is captured by three parameters. A unit normal vector can be characterized by two angles thus by (\ref{ref}) any reflection over a hyper-plane passing origin can be represented by two parameters.

The parameterization of linear isometry is the foundation of our statistical-based black-box attack, indeed the parameter space constitutes the sampling space. Further details of this subsection are discussed in supplementary materials.

\subsection{Thompson Sampling}\label{TS}
Thompson sampling (TS), also known as posterior sampling and probability matching, is a powerful sampling method that consecutively learns the prior probability of the parameters for sample distribution during the procedure. In this work, we adopt a Beta-Bernoulli finite bandit arm model \cite{russo2018tutorial} because of its efficiency in computation and conjugacy property.

Suppose there is an agent with $K$ actions, each action $k\in\{1,\cdots, K\}$ on the environment results in a reward one with probability $\theta_k\in[0,1]$ and zero with probability $1-\theta_k$. $(\theta_1,\cdots,\theta_K)$ are unknown to the agent but fixed over time. Suppose the agent has an independent prior belief for choosing $\theta_k$. Let the priors over $\theta_k$ to be the beta distribution with parameter $\alpha_k, \beta_k$. Then the prior probability density function for $\theta_k$ is
\begin{equation}
    p(\theta_k) = \frac{\Gamma(\alpha_k+\beta_k)}{\Gamma(\alpha)+\Gamma(\beta)}\theta_k^{\alpha_k-1}(1-\theta_k)^{\beta_k-1}
\end{equation}
where $\Gamma$ is the gamma function. Then after each observation from the environment the distribution of $\theta_k$ is updated according to Bayes' rule. By conjugacy property of beta distribution we can derive an update rule with simple form:
\begin{equation}
        (\alpha_k, \beta_k) \leftarrow \left\{
\begin{aligned}
&(\alpha_k, \beta_k)  &, k_t\neq k\\
&(\alpha_k, \beta_k)+(r_t, 1-r_t)  &, k_t=k 
\end{aligned}
\right.
\end{equation}
where $t\in\{1,\cdots,T\}$ denotes the time horizon, $k_t, r_t$ are the action and reward in time $t$. 

Later in our black-box attack settings, the action $k$ will be to sample a parameterized isometry in a local region corresponding to $k$. The reward is one if attack succeeds and zero otherwise. We assume the reward of attack follows Bernoulli distribution with success probability $\theta_k$, which is learnt by TS during the sampling.

\subsection{Restricted Isometry Property and Spectral Norm Penalty}\label{sec:ripsn}
The Restricted Isometry Property (RIP) was introduced by \cite{candes2005decoding} to develop theorems in the field of compressed sensing. RIP can describe the almost orthonormal system, which makes it a perfect mathematical tool to quantify the variation of an isometry to a restricted isometry.

For all $s$-sparse vectors $x\in\mathbb{R}^n$, i.e., $x$ has at most $s$ nonzero coordinates, matrix $A$ is said to satisfy $s$-restricted isometry with constant $\delta$ if
\begin{equation}\label{rip}
    (1-\delta)\Vert x\Vert^2\leq\Vert Ax\Vert^2\leq(1+\delta)\Vert x\Vert^2
\end{equation}
The smallest quantity of such $\delta$ is called the $s$-Restricted Isometry Constant (RIC) of $A$. We just consider the case $s=n$ since 1) there is no sparse constraint on vectors, 2) this allow us to impose orthogonality to $A$ from an RIP perspective \cite{bansal2018can}. The RIP condition in (\ref{rip}) is then
\begin{equation}
    \vert\frac{\Vert Ax\Vert^2}{\Vert x\Vert^2}-1\vert \leq \delta, \quad \forall x\in\mathbb{R}^n
\end{equation}
Let $\sigma(A)$ be the spectral norm of A, which is the largest singular value of $A$.
It can be easily shown
\begin{equation}
    \sigma(A^TA-I)=\sup_{x\in\mathbb{R}^n, x\neq 0}\vert\frac{\Vert Ax\Vert^2}{\Vert x\Vert^2}-1\vert 
\end{equation}
Hence, finding the $n$-RIC of $A$ is equivalent to computing $\sigma(A^TA-I)$. We call $\sigma(A^TA-I)$ the spectral norm penalty of $A$. It will be the core ingredient to help us develop the white-box attack in the next section.

\section{Attack Methods}
In this paper, we only consider adversarial attacks on classification models. Our approaches can be easily extended to other cases like segmentation models. 

\subsection{Point Cloud and Model Notations}
\noindent
\textbf{Point Cloud.}
Point cloud is a cluster of points in $\mathbb{R}^3$ that represents an object or a scene. Each point $p$ in a point cloud is described by its Cartesian coordinates $p=(x_p,y_p,z_p)^T\in\mathbb{R}^3$. Using a matrix notation, a point cloud $P$ with $m$ points is then described by $P=[p_1, p_2, \dots, p_m]^T\in\mathbb{R}^{m\times 3}$. \\

\noindent
\textbf{Deep 3D Point Cloud Model.}
A deep 3D point cloud model is a function 
$F_{\theta}(\cdot):\mathbb{R}^{m\times 3}\to \mathbb{T}^c$, where $\mathbb{T}^c=\{(a_1,\dots,a_c)\in\mathbb{R}^c_+: \sum_{j=1}^c a_j=1\}$ and $c$ is the total number of classes. We will omit $\theta$ later since in our analysis the model is fixed. Suppose $F$ is the full neural network with the softmax function and let $Z(P)=z\in\mathbb{R}^c$ be the output (called logit) except the softmax function, then $F(P)=\mathrm{softmax}(Z(P))$.

\subsection{Thompson Sampling Isometry (TSI) Attack}
The intuition of our isometry attack is to find an isometry on point cloud that misleads the model. For a given point cloud $P$, let $C(P)=\mathrm{argmax}_{j}F(P)_j$ be the classification result and $C^*(P)$ be the true label of $P$, then the \emph{isometry attack} is \begin{equation}\label{iso}
\begin{aligned}
 \mathrm{find}\quad &A\\
    s.t.\quad  & A\in \mathrm{Iso}(3)\\
          & \mathrm{argmax}_{j}F(PA^T)_j\neq C^*(P) 
\end{aligned}
\end{equation}

For better interpretation and conciseness, we will use the attack based on rotations as examples in this subsection. Following the results in subsection \ref{subsec:iso}, we can use three rotation angles $(\theta_x,\theta_y,\theta_z)\in [a,b]^3$ to construct an arbitrary rotation matrix of $\mathbb{R}^3$. It is not efficient to choose parameters from the uniform distribution $U([a,b]^3)$, let alone capturing any useful information during the sampling procedure. To address this issue, we utilize TS to design this attack. First we divide $[a,b]^3$ in to $d^3$ intervals of the form
\begin{equation}\label{interval}
    X_i\times Y_j\times Z_h
\end{equation}
for $(i,j,h)\in[d]^3:=\{1,\dots,d\}^3$, where
\begin{equation}\label{partition}
\begin{aligned}
    X_i&=[a+\frac{(i-1)(b-a)}{d},a+\frac{i(b-a)}{d}]\\
    Y_j&=[a+\frac{(j-1)(b-a)}{d},a+\frac{j(b-a)}{d}]\\
    Z_h&=[a+\frac{(h-1)(b-a)}{d},a+\frac{h(b-a)}{d}]
\end{aligned}
\end{equation}
 Applying an action $k=(i,j,h)$ means to sample $\theta_x\sim U(X_i),\theta_y\sim U(Y_j),\theta_z\sim U(Z_h)$ and use them to create a rotation matrix by (\ref{Euler}). Then we generate adversarial samples by this matrix and update parameters of TS. Our algorithm is as followed
\begin{algorithm}[H]
\footnotesize
 \caption{TSI$(F,\{P_1,\dots,P_N\},a,b,d,S)$}
 \begin{algorithmic}[1]\label{algo_ts}
 \renewcommand{\algorithmicrequire}{\textbf{Input:}}
 \renewcommand{\algorithmicensure}{\textbf{Output:}}
 \REQUIRE model $F$, point clouds $\{P_1,\dots,P_N\}$, parameters range $[a,b]$, number of intervals $d$, maximum sampling times $S$. 
 \ENSURE adversarial isometries $A_1^*,\dots,A_N^*$
  \STATE Initiate $\alpha_k\leftarrow 1,\beta_k\leftarrow 1,\forall k\in[d]^3$, where $k$ is corresponding to partition in (\ref{partition})
  \FOR {$n=1$ to $N$}
  \FOR {$s = 1$ to $S$}
  \STATE Sample $\hat{p}_k\sim\mathrm{beta}(\alpha_k,\beta_k),\forall k\in[d]^3$
  \STATE $k_s\leftarrow\mathrm{argmax}_{k}\hat{p}_k$
  \STATE Apply action $k_s$ and get an isometry matrix $A^{(s)}$
  \IF {$\mathrm{argmax}_{h}F(P_n(A^{(s)})^T)_h=C^*(P_n)$}
  \STATE $r_s\leftarrow 0$
  \ELSE
  \STATE $r_s\leftarrow 1$
  \ENDIF 
  \STATE $(\alpha_{k_s}, \beta_{k_s})\leftarrow (\alpha_{k_s}+ r_s, \beta_{k_s}+ 1-r_s)$  
  \ENDFOR
  \STATE $A_n^*\leftarrow\mathrm{argmin}_{A^{(s)}}F(P_n(A^{(s)})^T)_{C^*(P_n)}$
  \ENDFOR
 \end{algorithmic}
 \end{algorithm}
We assume the prior distribution of $p_s$ follows beta distribution $\mathrm{Beta}(\alpha_s,\beta_s)$ with mean $\frac{\alpha_s}{\alpha_s+\beta_s}$, which represents the attack success rate in area correspoding to $s$. In TSI, we can adopt an early stop if the attack succeeds before reaching maximal sampling times. 

\begin{figure}
\centering
\includegraphics[width=0.95\linewidth]{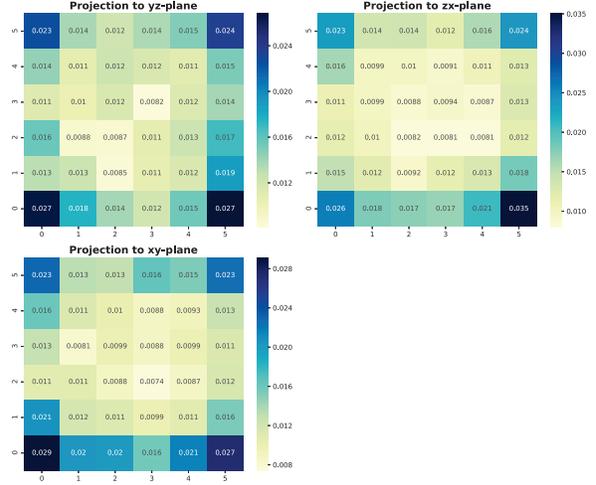}
\caption{Heat map of success rate in each region learnt by TS. The number of each area represents probability of successful attack if sampling angles from this region. We project the numbers along each axis by taking means, thus they can be seen as marginal probability of a successful attack.}\label{fig:tshot}
\end{figure}

To show the power of TS, in Figure \ref{fig:tshot} we present heat maps\footnote{In this attack evaluation, $d=6,[a,b]=[-\pi/16,\pi/16]$, $N=2000$ and $F$ is PointNet model trained by ModelNet40 data.} of $\frac{\alpha_s}{\alpha_s+\beta_s},s\in[d]^3$ projected to three planes. The heat map strongly testifies our insight that when the rotation angle gets closer to $0$, it is less likely to achieve a successful attack.

\subsection{Combined Targeted Restricted Isometry (CTRI) Attack}
 To further evaluate the robustness, we propose the \emph{restricted isometry attack} and formulate it as an optimization problem
\begin{equation}
    \begin{aligned}
    \min_{A\in \mathrm{GL}_3(\mathbb{R})} \quad &\sigma(A^TA-I)\\
         s.t. \quad & \mathrm{argmin}_jF(PA^T)_j\neq C^*(P)
    \end{aligned}
\end{equation}
Based on the results from subsection \ref{sec:ripsn}, the objective function is indeed the $3$-RIC of $A$. To impose a targeted attack, i.e., an attack that manipulates the classification result to be a targeted class $t$, we use the CW loss function introduced by \cite{carlini2017towards}
\begin{equation}
    g_t(P)=\max\{-\kappa,\max_{j\neq t}Z(P)_j-Z(P)_t\}
\end{equation}
where $\kappa\geq0$ is a chosen threshold. Then the Targeted Restricted Isometry (TRI) attack is formulated as
\begin{equation}\label{TRI}
    \begin{aligned}
     \min_{A\in \mathrm{GL}_3(\mathbb{R})} \quad \sigma(A^TA-I)+\lambda g_t(PA^T)
    \end{aligned}
\end{equation}
where $\lambda$ is a positive real number.

In practice we usually use Stochastic Gradient Descent (SGD) to iteratively solve optimization problems. To solve the problem in (\ref{TRI}), a good initial point is of vital importance. Fortunately, since for any matrix $A\in \mathrm{Iso}(3)$, $\sigma(A^TA-I)=0$, TSI actually provides a fine enough starting point for TRI attack. We now propose the Combined Targeted Restricted Isometry (CTRI) attack in Algorithm \ref{algo:ctri}.

\begin{algorithm}
\footnotesize
 \caption{CTRI Attack}
 \begin{algorithmic}[2]\label{algo:ctri}
 \renewcommand{\algorithmicrequire}{\textbf{Input:}}
 \renewcommand{\algorithmicensure}{\textbf{Output:}}
 \REQUIRE model $F$, point clouds $\{P_1,\dots,P_N\}$, parameters range $[a,b]$, number of intervals $d$, maximum iteration times $K$, maximum sample times $S$, learning step $\eta$, penalty coefficient $\lambda$, target class $t$ and $\kappa$.
 \ENSURE adversarial samples $P_1(A_1^*)^T,\dots,P_N(A_N^*)^T$
  \FOR {$n = 1$ to $N$}
  \STATE Set $A_n^{(0)}\leftarrow\mathrm{TSI}(F,\{P_n\},a,b,d,S)$, only initiate $(\alpha,\beta)$ in TS at the first time
  \STATE $k\leftarrow 0$
  \WHILE {$k < K$ and $F(P_n(A_n^{(k)})^T)=C^*(P_t)$}
  \STATE $A_n^{(k+1)}\leftarrow A_n^{(k)}-\eta\nabla_{A_n^{(k)}}[\sigma((A_n^{(k)})^TA_n^{(k)}-I)+\lambda g_t(P_n(A_n^{(k)})^T)]$ 
  \STATE $k\leftarrow k+1$
  \ENDWHILE
  \STATE $A_n^*\leftarrow A_n^{(k)}$
  \ENDFOR
 \end{algorithmic}
 \end{algorithm}

CTRI attack, enjoying both the merit of TS and the theoretical interpretation of RIP matrices, is a novel framework of white-box attack in 3D settings. By isometry parameterization, TS is capable to sample any isometry and explore the weakness of 3D model under different isometry transformations. Adjusting interval division parameters of TS gives us different scopes of examination, for example we can explore to what rotation range current 3D models are robust, which may provide insights on how to efficiently conduct data augmentation. As we will show in the experiment section, relaxation of isometry to RIP matrix produce a variety of adversarial samples that preserves the global geometry shape. Moreover, the spectral norm penalty, as a unified standard directly measured on transformation, makes the perturbation levels of different adversarial samples fairly comparable regardless the difference of data sets or models.

 
\section{Experiment Results and Analysis}
In this section, we evaluate the performance of our attacks on the following models: PointNet, PointNet++, DG-CNN and RS-CNN. We conduct our experiments on two data sets: ModelNet40 \cite{wu20153d} and part of ShapeNet \cite{yi2016scalable}. For conciseness and better interpretation, we will only use rotation matrices in this section. Experiments of other isometries such as reflections will be presented in supplementary materials, indeed attacks based on reflections performs even better under certain settings.

\subsection{Victim Models Setting}
We retrain the above models with normalized (to a unit sphere) and augmented data as the victim models. The augmentation includes random rotation about y axis, random scale, random translation and jitter \cite{qi2017pointnet}.  For ModelNet40, we use the official split of 9843 point clouds for training and 2468 for testing and attack, the sampling method is the same as those in \cite{qi2017pointnet}. For ShapeNet we use a 16 classes part with an official split to 12128 objects for training and 2874 for testing and attack. 2048 points are used for a single cloud in both data sets. The test accuracy of each retrained model is within 3\% of the best reported accuracy. 

In the following experiments, attacks will be exerted on these retrained models. Without extra statement, in one attack evaluation, we use 1) $N=2000$ point clouds; 2) parameter range $[a,b]=[-\pi,\pi]$ and division $d=4$ for TS. A point cloud that victim model fails to correctly classify is not counted in statistics of attack performance.
\subsection{TSI Attack}
We first evaluate TSI attack alone as an approach of black-box attack. Table \ref{tab:tsi} shows the results of $S=1,2,10$. 
\begin{table}
\footnotesize
\begin{center}
\begin{tabular}{l|c|c|c|c|c|c} 
\hline
&\multicolumn{3}{|c|}{ModelNet40} &\multicolumn{3}{|c}{ShapeNetPart}\\
\cline{2-7}
  & S=1 & S=2 & S=10 & S=1 & S=2 & S=10\\
\hline
PointNet & 92.69  & 96.92  & 98.28  & 83.10  & 92.61 & 98.11\\
PointNet++ &  83.74  &  91.31 & 95.66  & 79.32  & 89.57 & 92.96\\
DG-CNN & 79.57 & 89.81 & 97.33 & 73.69  &87.17  & 96.82\\
RS-CNN  & 85.68  & 94.19 & 96.81 & 76.46  &90.01 & 97.41\\
\hline
\end{tabular}
\end{center}
\caption{Attack success rates (\%) of TSI with different maximum sampling times $S$.}\label{tab:tsi}
\end{table}
 As shown in the table, with maximum sampling times increasing to $10$, TS achieves an attack success rate over 95\% on ModelNet40, revealing the vulnerability of current 3D models under isometry. 

 Current black-box attacks in 2D are mainly through adversarial transferability \cite{papernot2016transferability}, heuristic methods \cite{alzantot2019genattack} and estimation of gradients \cite{chen2017zoo}. These methods may not work in 3D settings due to different structure of data, for example, \cite{xiang2019generating} finds that transferability is not obvious in 3D adversarial attacks.  In comparison, TSI performs exceptionally well as a black-box attack. 

The reason why we adopt TS is that we believe for different rotation angles the impact on victim model is different. For example, if the rotation angle is very close to $0$, attack might not work, but if it is close to $\pi/4$, attack might easily succeed. This insight is strongly confirmed by heat maps in Figure \ref{fig:tshot}. By understanding isometry properties of data set and model, TSI approach also shades light on how to conduct more effective and efficient data augmentation. We further guess a more delicate designed Thompson Sampling model such as an infinite-arm model might work better. Moreover, it might also perform well to generate different parameterized transformation other than isometry. We leave them for future work.

\subsection{CTRI Attack}
We then evaluate CTRI, where the target is chosen to be the label with second largest logit predicted by victim models, and $\kappa=0$ is adopted. In order to find a good enough initial point, we use $S=50$ in TSI. The attack success rate is presented in Table \ref{tab_ctri} for maximum gradient iteration times $K=7, 50, 1000$. We adopt an early stop here, i.e., when TSI or CTRI succeeds before reaching maximum iteration times, we move to the next sample. We also show the statistics of $\sigma(A^TA-I)$ in Table \ref{tab_stat}, \ref{tab_stat1} for restricted isometry $A$ that successfully fools the victim models.
\begin{table}
\footnotesize
\begin{center}
\begin{tabular}{l|c|c|c|c|c|c} 
\hline
 &\multicolumn{3}{|c|}{ModelNet40} &\multicolumn{3}{|c}{ShapeNetPart}\\
\cline{2-7}
 & K=7 &K=50 &K=1000 &K=7 &K=50 &K=1000\\
\hline
PointNet & 99.44 & 99.66 & 100.00 & 99.38 & 99.69 & 100.00\\
PointNet++ & 97.93 & 98.11 & 98.88 & 96.19 & 96.71 & 99.34\\
DG-CNN & 97.99 & 98.28 & 99.45 & 98.26 & 99.08 & 99.95\\
RS-CNN  & 98.21 & 98.38 & 99.33 & 98.23 & 98.84 & 99.95\\
\hline
\end{tabular}
\end{center}
\caption{Attack success rates (\%) of CTRI with different $K$ , same $\lambda=0.001$ and learning step $\eta=0.0005$. Here we use $S=50$ in TSI.}\label{tab_ctri}
\end{table}

\begin{table}
\footnotesize
\begin{center}
\begin{tabular}{l|c|c|c|c|c} 
\hline
  &Max & Mean & Var & Mean* & Var*\\
\hline
PointNet & 0.064  & 8E-5 & 4E-6 & 0.033 & 3E-4\\
PointNet++ &  0.055  & 1E-4 & 3E-6 & 0.006 & 9E-5\\
DG-CNN & 0.066 & 9E-5 & 4E-6 & 0.020 & 5E-4\\
RS-CNN  & 0.068  & 6E-5 & 3E-6 & 0.025 & 5E-4 \\
\hline
\end{tabular}
\end{center}
\caption{Statistics of $\sigma(A^TA-I)$ for the second column of table \ref{tab_ctri} (ModelNet40, $K=50$). Since we adopt early stop, $\sigma(A^TA-I)$ will be zero if TSI succeeds. 'Mean*' and 'Var*' are the mean and variance over nonzero ones. They better reflect difference caused by TRI alone.}\label{tab_stat}
\end{table}

\begin{table}
\footnotesize
\begin{center}
\begin{tabular}{l|c|c|c|c|c} 
\hline
  &Max & Mean & Var & Mean* & Var*\\
\hline
PointNet & 1.11  & 0.002 & 0.001 & 0.331 & 0.107\\
PointNet++ &  1.00  & 0.004 & 0.003 & 0.114 & 0.063\\
DG-CNN & 0.99  & 0.004 & 0.003 & 0.276 & 0.091\\
RS-CNN  & 0.84  & 0.003 & 0.002 & 0.318 & 0.073\\
\hline
\end{tabular}
\end{center}
\caption{Statistics of $\sigma(A^TA-I)$ for the third column of table \ref{tab_ctri} (ModelNet40, $K=1000$).}\label{tab_stat1}
\end{table}

\begin{figure}
\begin{center}
   \includegraphics[width=1\linewidth]{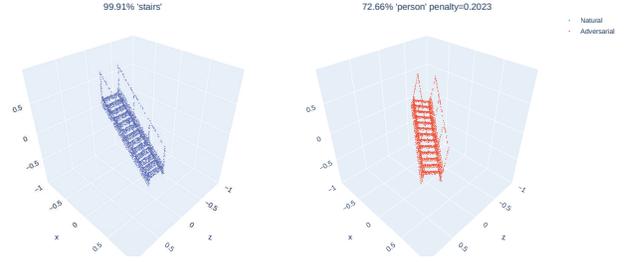}
\end{center}
   \caption{Illustration of CTRI on PointNet trained by ModelNet40 data. The object in blue is classified as 'stairs' with confidence 99.91\%, while the grey object generated by CTRI is predicted as 'person' with confidence 72.66\%. Spectral norm penalty $\sigma(A^TA-I)=0.2033$.}
\label{fig:ctri}
\end{figure}

 Visualization of TSI and CTRI adversarial examples are shown in Figure \ref{fig_tsi} and \ref{fig:ctri}. Figure \ref{fig_tsi} shows misclassification of a rotated highway cone, which may cause safety issues for self-driving cars. Figure \ref{fig:ctri} consists of natural and adversarial samples of a stair, where the global shape is well preserved in the adversarial sample. More examples 
 are demonstrated in the supplementary materials.

\subsection{Transferability}
Transferability is a commonly-seen property in 2D adversarial samples \cite{papernot2016transferability}. It refers to the case that an adversarial sample generate from one model can mislead another different model. However, in deep 3D models the adversarial samples proposed by previous works do not have strong transferability \cite{xiang2019generating,yang2019adversarial}.

On the contrary, we find that adversarial samples generated by CTRI attacks have strong transferability, as shown in Table \ref{tab_trans}. In fact, a transfer attack should perform at least not worse than a random generated isometry, i.e., TSI attack when $S=1$. Such high rates indicate again that these models are vulnerable to isometry transformations. Another reason why our approach has stronger transferability than that of others is that, CTRI tries to explore the global geometry structure, while others depend much more on the property of specific model such as critical points.       

\begin{table}
\footnotesize
\begin{center}
\begin{tabular}{l|c|c|c|c} 
\hline
 & PointNet & PointNet++ & DG-CNN & RS-CNN\\ 
\hline
PointNet & /  & 86.61  & 80.62 & 85.40 \\
PointNet++ &  95.76  & / & 85.68 & 92.71\\
DG-CNN & 95.53  & 93.34 & / & 93.11 \\
RS-CNN  & 95.04  & 92.58 & 84.59 & / \\
\hline
\end{tabular}
\end{center}
\caption{Transfer attack success rate (\%) of CTRI with $S=50, K=1000$. The adversarial sample is generated from the model corresponding to the row and tested in model corresponding to the column. }\label{tab_trans}
\end{table}

\subsection{Attack of Small Rotation Angles}\label{subsec:small}
Instead of random sampling rotation angles from $[-\pi,\pi]$, in this subsection we examine TSI and CTRI attack with a smaller range, say $[-\epsilon, \epsilon]$. The result is shown in Table \ref{tab:small}. TSI attack maintains a success rate of $69.529\%$ when the range shrinks to $[-\pi/16,\pi/16]$. It only fails when the range reduces to an imperceptible level ($\pi/64=2.81^{\circ}$), indicating the model is just robust to relatively small range of rotations. Meanwhile CTRI attack succeeds easily with rates over $95\%$ and means of penalties less than $0.3$. 

This experiment also provides useful information for efficient data augmentation, which we will discuss in the following subsection. 

\begin{table*}
\footnotesize
\begin{center}
\begin{tabular}{l|c|c|c|c|c|c|c} 
\hline
  & CTRI (\%) & TSI (\%) & Max & Mean & Var & Mean* & Var* \\ 
\hline
$[-\pi/2\ \ ,\pi/2 \ \ ]$ &100.000 & 99.335 &1.012 & 0.001 & 8E-4 & 0.201 & 0.076  \\
$[-\pi/4\ \ ,\pi/4 \ \ ]$ &99.944 & 99.386 &0.443 & 0.001 & 4E-4 & 0.197 & 0.024  \\
$[-\pi/8\ \ ,\pi/8\ \ ]$ & 99.944 & 98.153 & 1.460 & 0.003 & 0.003 & 0.214 & 0.101 \\
$[-\pi/16, \pi/16]$ &99.776 & 69.529 & 1.406 & 0.037 & 0.010 & 0.139 & 0.021  \\
$[-\pi/32,\pi/32]$ & 98.222 & 18.000 & 1.833 & 0.178 & 0.047 & 0.246 & 0.047 \\
$[-\pi/64,\pi/64]$ & 95.536 & 5.469 & 2.605 & 0.287 & 0.083 & 0.356 & 0.080 \\
\hline
\end{tabular}
\end{center}
\caption{Attack evaluation of small rotation angles on ModelNet40 data and PointNet model. The first two columns are the attack success rate of iterative CTRI and TSI with $K=1000, S=100$, the rest are statistics for $\sigma(A^TA-I)$. Other parameters are the same as those in Table \ref{tab_ctri}.}\label{tab:small}
\end{table*}

\subsection{On Defense: the Price of Robustness}
Data augmentation usually slightly lowers down the accuracy but helps improve the robustness \cite{li2018pointcnn}. However, our experiments show current data augmentation methods in 3D (random rotation about y, random scale and translation, jitter) are not enough for defense against our attacks. 

\begin{figure}
\centering
\includegraphics[width=1\linewidth]{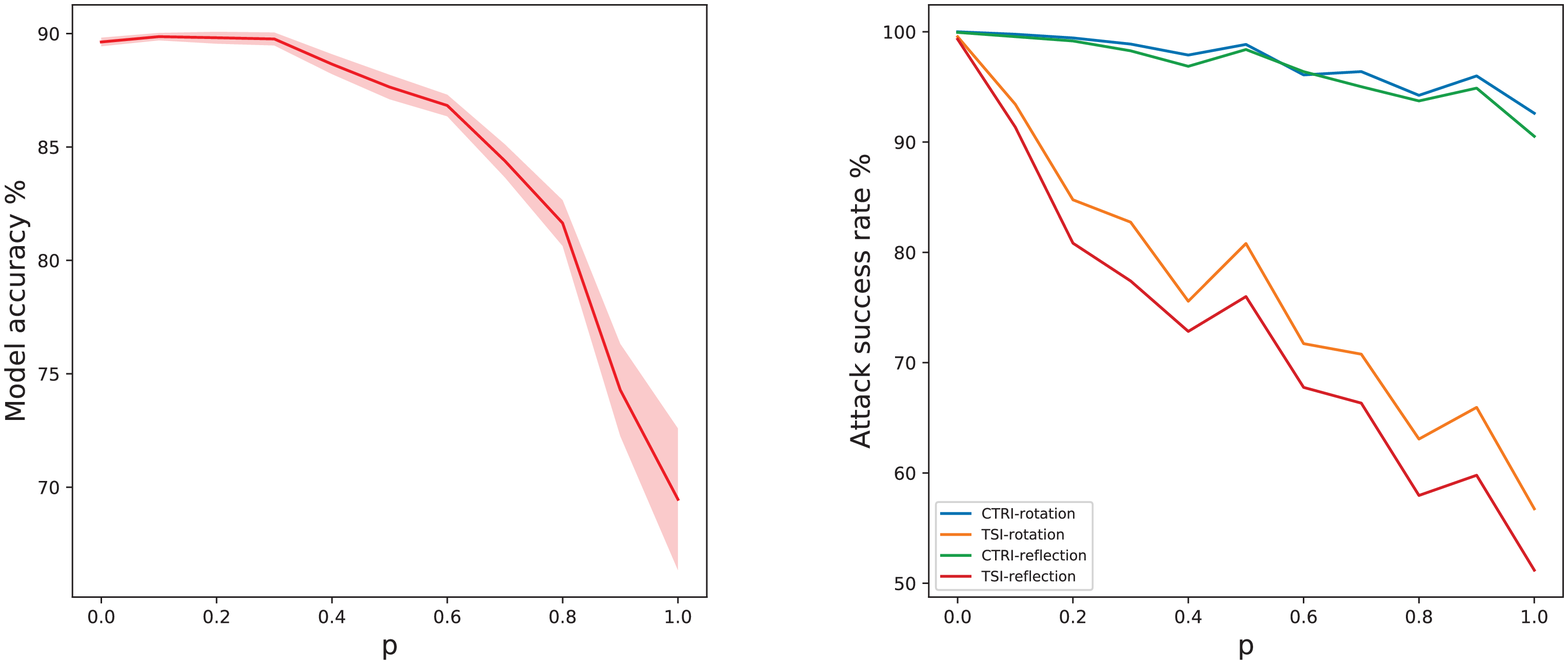}
\caption{When $p$ goes up, the right figure shows the attack success rates decrease, while the left figure suggests model accuracy also goes down and the variance increases. Thus, there is a trade-off between model accuracy and robustness when data augmentation approach is considered.}\label{fig_price}
\end{figure}

To better understand the relation between robustness and accuracy from a data augmentation approach, we design an experiment: 1) First, we propose an augmentation method called \emph{random $p$-rotation}. It is to randomly rotate the data by (\ref{Euler}) with probability $p$. 2) Then we train PointNet model with random $p$-rotation ModelNet40 data 15 times for different $p\in\{0,0.1,\dots,0.9,1.0\}$. 3) Finally, we randomly choose a victim model for each $p$, and launch both rotation and reflection based TSI and CTRI attacks (same parameters as those in Table \ref{tab_ctri}, $K=1000$) on these models. The result is demonstrated in Figure \ref{fig_price}.

Clearly, accuracy varies a lot with different $p$. The mean accuracy goes down to 69.47\% when $p=1$. TSI attack success rates decrease as $p$ increases, but CTRI success rates still remain in a high level. Thus, the point is how to find an efficient way of data augmentation without jeopardizing the accuracy too much. As shown in Figure \ref{fig_price}, roughly when $p\in[0.1,0.3]$, the model accuracy remains around $90\%$ while TSI attack success rates reduce to around $80\%$. Hence, random rotation with a low probability is a relatively efficient way to defense against TSI attack. 

Another efficient data augmentation method could be to randomly rotate the data by several fixed angles, say $\epsilon,2\epsilon,\dots,k\epsilon$, where $\epsilon$ is chosen from subsection \ref{subsec:small}. The intuition is that since current models are robust to small rotation angles, we can take $\epsilon$ as the minimum step length to rotate the data. This approach is left for future work.

 We note that 3D augmentation should draw more attention, since many data sets in 3D are often not so adequate for training large models. In particular, simple augmentations are not sufficient under adversarial attacks. For one thing, data augmentation often brings large variance, which decreases the performance of models \cite{Lan2019}. For another, augmentations without understanding of attacks are less efficient, and can lead to computational difficulties \cite{Bubeck2019}. An efficient data augmentation may be able to defense against TSI without too much cost, but against CTRI attack a lot more computational power might be required, for example, by adversarial training \cite{madry2017towards}.  
 

\section{Conclusion}
We propose a black-box attack and a white-box attack in 3D adversarial settings, which is the first work to show the extreme vulnerability of current 3D deep learning model under isometry transformations. On the one hand, there are intriguing properties of our adversarial samples, such as strong transferability. On the other hand, our approaches point out promising directions for future researches in 3D adversarial learning: 1) TS shows great potential in developing black-box attack, which can also be further exploited in data augmentation. 2) CTRI achieves satisfying performance even under small rotation angles, while the geometry property of adversarial samples is well preserved. Therefore, further researches on defense against TSI and CTRI may hugely improve isometry robustness of 3D deep learning models. 

\section{Acknowledgement}
The first author would like to thank Wang Qingyu for the help in plotting, Li Xinke for discussion, and Pan Binbin for proof reading. The work is supported under Grant NRF-RSS2016-004,
NSF of Jiangsu Province BK20181259, NSFC 11871269 and NSFC 71673130.

{\small
\bibliographystyle{ieee_fullname}
\bibliography{egbib}
}

\end{document}